\pdfoutput=1

\documentclass[11pt]{article}

\usepackage[]{acl}
\usepackage{subcaption} 
\usepackage{listings}
\usepackage{xcolor}

\usepackage{times}
\usepackage{latexsym}
\usepackage[most]{tcolorbox}
\usepackage{colortbl} 
\usepackage{xcolor}   
\usepackage{soul}
\usepackage{multirow}
\usepackage{float} 
\usepackage{tabularx} 
\usepackage{amsmath}
\usepackage{amssymb}
\usepackage{caption}
\usepackage{graphicx}
\usepackage{wrapfig}
\usepackage{float}
\usepackage{algorithm}
\usepackage{booktabs}
\usepackage{appendix}
\usepackage{xcolor}
\usepackage{pifont}
\newcommand{\cmark}{\textcolor{green!60!black}{\ding{51}}}  
\newcommand{\xmark}{\textcolor{red}{\ding{55}}}              
\usepackage{xcolor}
\usepackage{graphicx}
\usepackage{booktabs}
\usepackage{multirow}
\usepackage{enumitem}
\usepackage{listings}
\usepackage{xcolor}
\usepackage{graphicx}
\usepackage{url}
\usepackage[utf8]{inputenc}
\usepackage{graphicx}
\usepackage{caption}
\usepackage{amsmath}
\usepackage{amsthm}
\usepackage{booktabs}
\usepackage{algorithm}
\usepackage{algorithmic}
\usepackage[switch]{lineno}
\usepackage[T1]{fontenc}

\usepackage{microtype}

\usepackage{inconsolata}

\usepackage{graphicx}

\title{Not All Options Are Created Equal: Textual Option Weighting for Token-Efficient LLM-Based Knowledge Tracing}

\author{
JongWoo Kim\textsuperscript{1,*} \quad
SeongYeub Chu\textsuperscript{1,*} \quad
Bryan Wong\textsuperscript{1} \quad
Mun Yi\textsuperscript{1,\dag} \\
\textsuperscript{1}KAIST, Daejeon, Republic of Korea \\
\texttt{\{gsds4885, chseye7, bryan.wong, munyi\}@kaist.ac.kr}
}

\begin{document}
\maketitle

\begin{abstract}
Large Language Models (LLMs) have recently emerged as promising tools for knowledge tracing due to their strong reasoning and generalization abilities. While recent LLM-based KT methods have introduced new prompt formats, they struggle to reflect all histories of example learners within a single prompt during in-context learning (ICL), leading to limited scalability and high computational cost under token constraints. In this work, we present \textit{LLM-based Option weighted Knowledge Tracing (LOKT)}, a simple yet effective LLM-based KT framework that encodes the interaction histories of example learners in context as \textit{textual categorical option weights (TCOW)}. These are semantic labels (e.g., “inadequate”) assigned to the options selected by learners when answering questions, helping understand LLM. Experiments on multiple-choice datasets show that LOKT outperforms existing non LLM- and LLM-based KT models in both cold-start and warm-shot settings. Moreover, LOKT enables scalable and cost-efficient inference, performing strongly even under strict token constraints. Our code is available at \href{https://anonymous.4open.science/r/LOKT_model-3233}{https://anonymous.4open.science/r/LOKT\_model-3233}.
\end{abstract}


\section{Introduction}



Knowledge Tracing (KT) is a core method in learning analytics that aims to estimate and track a learner’s evolving understanding of specific knowledge components (KCs) over time.  It models how a learner’s knowledge state changes with each learning interaction, such as answering a question, and uses this to predict future performance \cite{Piech2015, Xia2019}. However, its effectiveness drops in cold-start scenarios where interaction data is limited \cite{fu2024sinkt, zhao2020cold}.

\begin{figure}[t]
    \centering 
    \includegraphics[width=\linewidth]{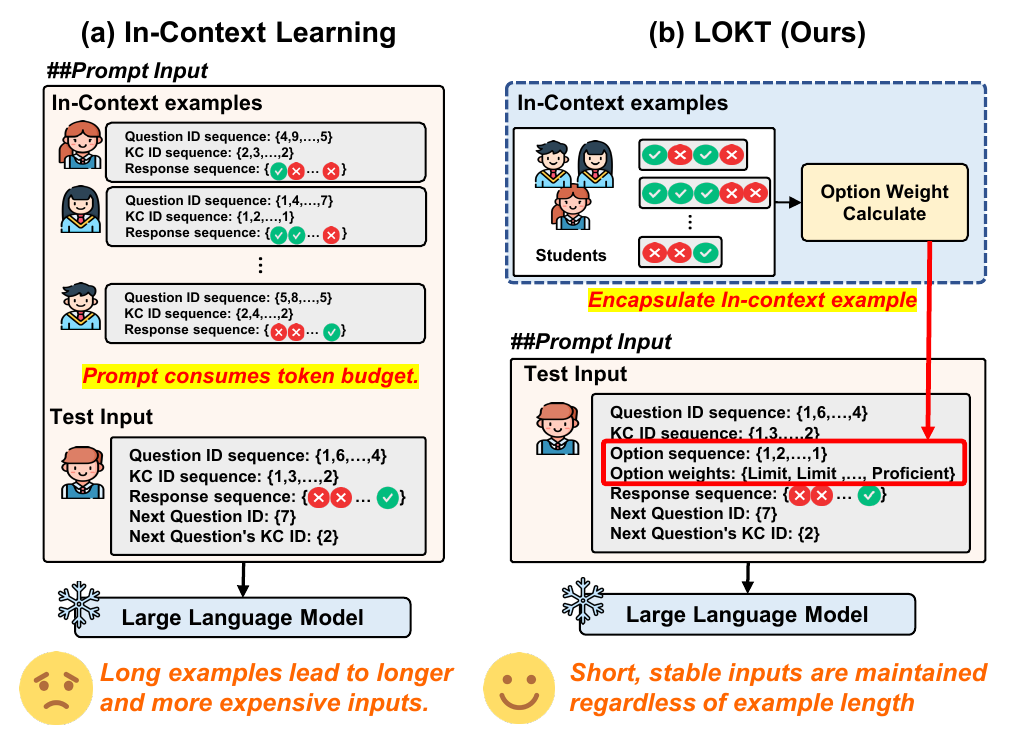}
     
    \caption{Comparison between the ICL approach and LOKT. (a) ICL requires including complete learner interaction histories in prompts, resulting in increased token usage, while (b) LOKT compresses information through textual option weights, maintaining constant token usage while achieving superior performance.}
    \label{fig:option_vs_icl}
      
\end{figure}

To tackle the bottleneck, there have been increasing efforts to leverage Large Language Models (LLMs), which possess broad prior knowledge and strong reasoning capabilities \cite{santoso2024large, LLMforKT2023}. Most existing approaches rely on either fine-tuning (FT) LLMs using learner interaction data \cite{jung2024, LLMforKT2023} or in-context learning (ICL) \cite{li2024explainable}. Particularly, the ICL-based approach has recently gained attentions owing to its flexibility and practicality by providing prompts without updating model parameters. As shown in Figure~\ref{fig:option_vs_icl}-(a), this method predicts how the test student will answer the next question by looking at a few example sequences from other students, which are provided in the prompt as references.

\begin{figure}[t]
    \centering
    \includegraphics[width=\linewidth]{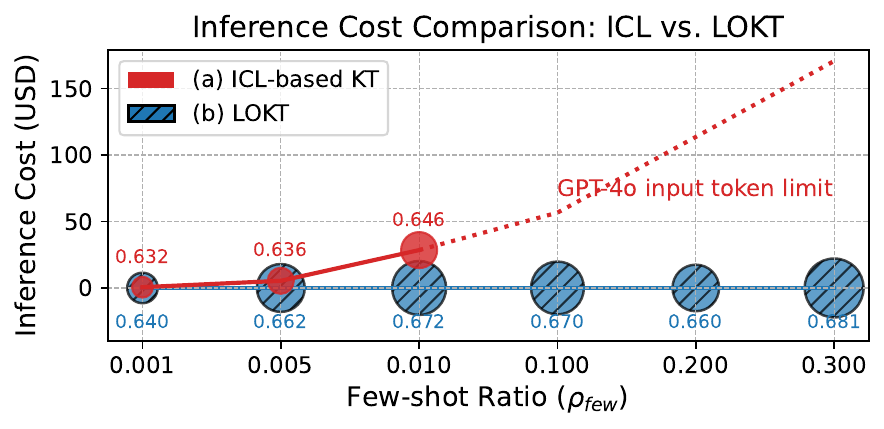}
    
    \caption{Inference cost and performance across few-shot ratios for (a) ICL-based KT and (b) LOKT. ICL incurs rising cost and hits GPT-4o’s token limit as few-shot size increases. In contrast, LOKT maintains fixed cost while improving with more examples.}
    \label{fig:icl}
     
\end{figure}

While ICL-based KT has shown promising performance in few-shot prompt settings, it raises a key question: \textbf{Is this approach truly effective for KT tasks, where each example consists of long interaction sequences?} To investigate this, we examined the cost, scalability, and performance of ICL-based KT across various few-shot configurations (Figure~\ref{fig:icl}). Including more examples in the prompt increases token usage and inference cost, while reducing them due to token budget constraints degrades performance. These patterns, consistent with findings in other domains \cite{jiang2023longllmlingua, li2024explainable, han2024token, liu2023tcra}, highlight the structural limitations of ICL for long-context KT scenarios. 

To address these limitations, we propose a novel method called \textbf{LOKT}(LLM-based Option weighted Knowledge Tracing). As shown in Figure~\ref{fig:option_vs_icl} (b), LOKT generates an encapsulated representation of learner behavior by using option weights, instead of inserting full interaction histories into prompts as in prior ICL-based work. This approach allows LOKT to maintain a fixed input length regardless of the number of learner interactions. Option weights quantify learners' current knowledge states represented in the selected option in multiple-choice question (MCQ) responses. These weights support systematic analysis of learner understanding or misconception based on option-level selections. However, their continuous values (ranging from -1 to 1 \cite{huang2024pull,an2022no}) are difficult for LLMs to interpret \cite{santoso2024large}. To address this, we propose textual categorical option weights (TCOW) which transforms the continuous weights into textual categories such as \textit{Inadequate}, \textit{Limited}, and \textit{Proficient} (see Figure~\ref{fig:option_vs_icl} b). This facilitates LLMs' comprehension of the information lying in selected options.


We conduct experiments on five multiple-choice knowledge tracing datasets. To assess the effectiveness of our approach, we compare LOKT with both LLM-based baselines and conventional KT models under cold-start condition. Our findings show that LOKT improves LLMs' knowledge tracing performance with shorter prompts which save costs. Our contributions are as follows:
 
\begin{itemize}
    \item We introduce LOKT, a method that leverages option weights to distill long-context information into a compact sequence, thereby enhancing KT performance in a cost-effective and efficient manner, particularly in cold-start scenarios.
    
    \item We propose TCOW, which further transforms the continuous option weights into textual categories, enhancing LLMs' interpretation of the option information during KT.
     
    \item Experiments on five multiple-choice KT datasets show that LOKT achieves higher accuracy and stability in diverse cold scenarios and warm scenarios than previous KT and LLM-based methods.
\end{itemize}

\begin{figure*}
    \centering
    \includegraphics[width=1\linewidth]{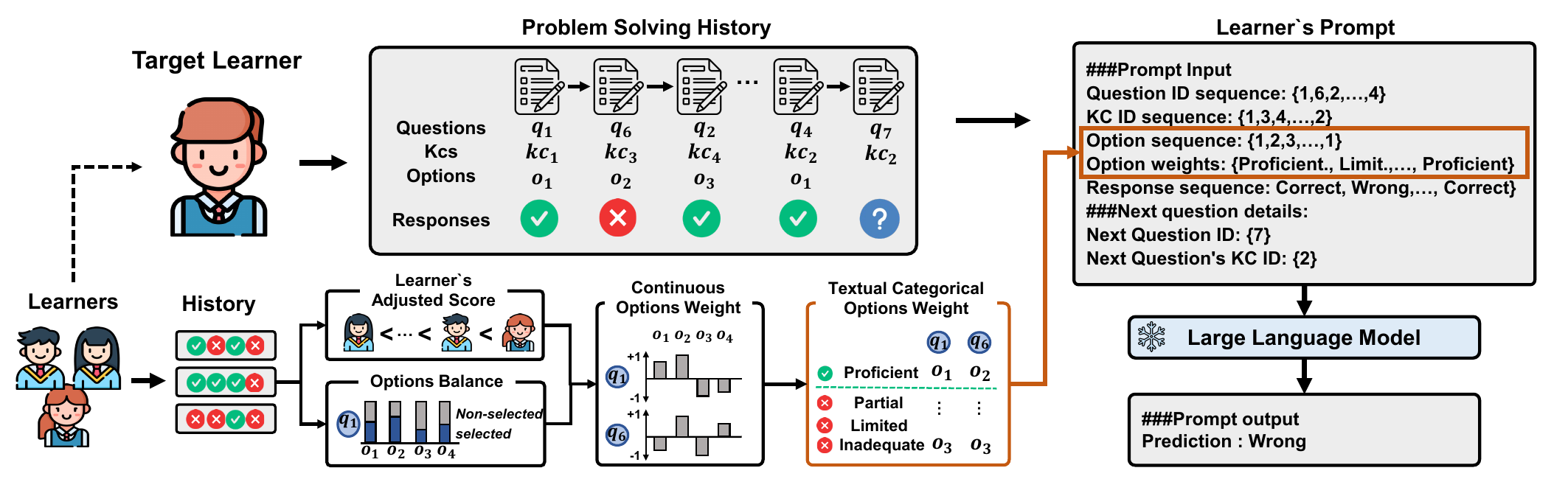}
    \caption{Option weights are first computed from peer learners’ interaction data, incorporating learners’ adjusted scores and option selection patterns. These option weights are converted into textual form that the LLM can better understand. By using these option weights, we encapsulate the information of in-context example learners, enabling effective representation without directly including the full interaction history in the prompt.}
    \label{fig1}

\end{figure*}

\section{Related Work}

\subsection{Knowledge Tracing}

Knowledge tracing (KT) aims to model and predict a learner’s evolving mastery of KCs based on their past interactions with educational content. KT originated from statistical methods such as maximum likelihood estimation (MLE), item response theory (IRT) \cite{Lord1980}, and Bayesian knowledge tracing (BKT) \cite{Corbett1995}. Since the evolution of neural networks, models like DKT \cite{Piech2015} and DKT+ \cite{Chen2020} leveraged RNNs and regularization to capture temporal patterns and improve stability. Attention-based models, including KQN \cite{lee2019knowledge}, SAKT \cite{Xia2019}, and their variants \cite{Kim2020, Hu2021}, enhanced prediction of learners' knowledge state by focusing on salient student-exercise interactions. More recent methods like ReKT \cite{shen2024revisiting} and ExtraKT \cite{li2024extending} integrate contextual signals to accurately trace pupils' knowledge states. However, these models remain limited in cold-start settings with limited  interaction history \cite{LLMforKT2023}. To tackle the bottleneck, we newly incorporate textual option weights into LLMs' knowledge tracing.

\subsection{LLM based Knowledge Tracing}

LLMs have recently emerged as strong alternatives for KT, leveraging broad prior knowledge and reasoning to infer learners' states from limited data \cite{Brown2020, LLMforKT2023}. Most approaches focus on prompt engineering rather than introducing new model architectures, typically relying on fine-tuned LLMs. Neshaei et al.\ use prompts summarizing KC correctness rates \cite{neshaei2024towards}, while CLST suggests a [KC, Response] pair sequence as a prompt, framing KT as a language modeling task \cite{jung2024}. Fine-tuning poses practical limitations in real-world settings, such as requiring separate training for distinct datasets. Consequently, inference-only methods such as ICL have gained attention by embedding learner histories into prompt-templates without updating model parameters \cite{li2024explainable}. However, its scalability is still limited due to prompt length, token costs, and other challenges such as context management and inference latency \cite{Zhao2023}.To address these challenges, we propose LOKT, an efficient prompt-engineering method that uses textual option-weight summaries for efficient KT.
 

\section{Methodology}

\subsection{Overview}

LOKT introduces a structured prompting framework that compresses learner interactions into semantically meaningful representations to address the limitations of in-context learning in knowledge tracing. Figure~\ref{fig1} illustrates the overall architecture. We derive continuous option weights from peer learners, transform them into textual form, and incorporate them into the learner’s interaction sequence to construct structured prompts.

\subsection{Problem Definition}

\noindent\textbf{\textit{Knowledge Tracing.}}
KT aims to model and predict a learner’s evolving mastery of KCs based on their past interactions. It not only predicts whether a learner will correctly answer the next question but also captures temporal patterns in their learning behavior. Each interaction \( x_{t,i} \) at time step \( t \) for learner \( i \) includes the question \( q_{t,i} \), its associated knowledge components \( c_{t,i} \), and the response label \( y_{t,i} \). The goal is to estimate the correctness of the next response \( \hat{y}_{t+1,i} \), computed as:

\begin{equation}
\hat{y}_{t+1,i} = \arg\max_{\omega} P(\omega \mid q_{t,i}, c_{t,i})
\end{equation}

Here, \( \omega \) is the predicted label indicating whether the learner will answer correctly. This formulation models KT as a sequential prediction task.
 
\noindent\textbf{\textit{Few-shot Cold-start Setting.} } 
This setting evaluates the model's generalization ability when only a limited number of learners are available. Non-LLM models follow a few-shot training setup, where the proportion of learners used for training is determined by \( \rho_{\text{few}} \). In contrast, LLM-based models perform inference without any parameter updates and rely on ICL, where a small number of exemplar learners corresponding to \( \rho_{\text{few}} \) are included in the prompt, subject to token length constraints. Unlike general NLP tasks where few-shot size is defined by the number of exemplars (e.g., 2, 4, or 8), few-shot learning in KT is typically defined based on the proportion of training data.

\subsection{Learning from Option Information}

LLM-based KT models often simplify responses into correct or wrong classifications, overlooking detailed insights from learners’ option choices. Option weights address this gap by analyzing misconceptions in wrong responses, providing a more granular assessment of knowledge \cite{ghosh,an2022no,huang2024pull}. However, continuous option weights, while capturing quantitative differences, pose challenges in integration with LLMs as the range of weights varies depending on questions where the same option weights of different exercise can indicate different insights. It can cause confusion for LLMs~\cite{santoso2024large}. To overcome this, we replace these numerical weights into text-based categorical values, enhancing the LLMs' ability to interpret learners' proficiencies and misconceptions effectively.

\subsubsection{Continuous Option Weight}
 
We adopt the method from \cite{huang2024pull}, which computes continuous option weights in two steps: estimating learner proficiency and aggregating these estimates over option choices. Detailed calculations are provided in the Appendix ~\ref{optionweight}. 

\noindent\textbf{\textit{Learner Proficiency Measurement.}} For option weights calculation, we first measure each learners' proficiency based on their average correct rate, adjusted by exercise difficulty to account for varying problem-solving logs and question sets. The difficulty of a question is defined as \( d_q = \frac{\text{correct}_q}{\text{count}_q} \), where \(\text{count}_q\) denotes the number of attempts on question \(q\), and \(\text{correct}_q\) denotes the number of correct answers. Subsequently, each learner’s score \( \bar{x}_i \) is calculated by subtracting the average difficulty of the questions they attempted from their total number of correct answers:

\begin{equation}
    \bar{x}_i = \frac{N_i - \sum_{q \in Q_i} d_q}{|Q_i|}
\end{equation}

Here, \( i \) denotes a learner, \( N_i \) is the number of correctly answered questions, and \( Q_i \) is the set of questions attempted by learner \( i \). This adjustment increases the score for correctly answering difficult questions and decreases it for wrong answers to easier ones. Then, we normalize the adjusted scores using the global mean (\(\bar{x}\)) and standard deviation (\(S_x\)) computed over all learners \( I \).

\noindent\textbf{\textit{Option Weight Calculation.}} 
To compute option weights that reflect the diagnostic relevance of each choice, we analyze how option selection correlates with learner proficiency. The objective is to assign higher weights to options that are indicative of either understanding or common misconception.

For a given question \( q \) and option \( o \), we compute the selection ratio as \( C^o_q = \frac{c^o_q}{c^o_q + nc^o_q} \), where \( c^o_q \) and \( nc^o_q \) denote the number of learners who selected and did not select the option, respectively. The non-selection ratio is defined as \( NC^o_q = 1 - C^o_q \). We then divide learners into two groups and compute their average adjusted scores:

\begin{equation}
\bar{x}_{c^o_q} = \frac{\sum_{i \in I^o_q} \bar{x}_i}{|C^o_q|}, \quad \bar{x}_{nc^o_q} = \frac{\sum_{i \in NI^o_q} \bar{x}_i}{|NC^o_q|}
\end{equation}

where \( \bar{x}_i \) is the adjusted proficiency score of learner \( i \), \( I^o_q \) denotes the set of learners who selected option \( o \) for question \( q \), and \( NI^o_q \) denotes the set of learners who did not select option \( o \). The difference between these two averages, \( \bar{x}_{c^o_q} - \bar{x}_{nc^o_q} \), reflects how strongly the option's selection correlates with learner proficiency. Using these statistics, we define the continuous option weight as:

\begin{equation}
    w^o_q = \frac{\bar{x}_{c^o_q}-\bar{x}_{nc^o_q}}{S_x} \times \sqrt{C^o_q \cdot NC^o_q}
\end{equation}

\subsubsection{Textual Categorical Option Weight}

For representation of the relative importance of options, we sort the weights for question \( q \) as \( w^{(1)}_q \ge w^{(2)}_q \ge \cdots \ge w^{(n_q)}_q \), which provides ordinal position but lacks semantic meaning, limiting its utility for language models. To enhance interpretability and better align with LLM capabilities, we convert these ranks into semantically meaningful textual categories. Each option is assigned a category based on its rank: the highest (\( w^{(1)}_q\)) is \textit{Proficient}, then \textit{Partial}, \textit{Limited}, and the lowest (\( w^{(n_q)}_q\)) is \textit{Inadequate}. based on its relative weight. This transformation supports LLMs' interpreting learners' understanding, where \textit{Proficient} reflects strong mastery and \textit{Inadequate} suggests misconceptions. When included in prompts, these descriptors offer semantically rich signals that help the LLM reason about learner behavior. Some options may lack sufficient observations to compute stable weights; in such cases, we set the weight to \texttt{NaN}. For completeness, mappings for questions with more than four options are provided in Appendix~\ref{multiple_choice}.

\subsection{Prompt Construction for KT}













%
%

To predict whether a learner will answer correctly at time \(t+1\), we encode their interaction history up to time \(t\) into a structured prompt. To guide LLMs in capturing problem-solving components, namely question, KC, options, option weights, and response, we separately provided the sequence of each component to LLMs. The prompt includes the target learner's problem-solving history, along with their selected options and the corresponding TCOW representations. We provided question ID corresponding KC ID at time \(t+1\) at the end of the prompt-template. Option weights are computed from the interaction of learners and exercises in few-shot examples, then transformed into TCOWs associated with the target learner's chosen options. The final prompt is fed into a frozen LLM (e.g., GPT-4o) to predict the learner’s next response as either \texttt{correct} or \texttt{wrong}. An example of this prompt structure is shown in Appendi \ref{prompt_1} .

\section{Experiment}

This section presents the experimental setup and results that validate the effectiveness of the proposed methodology. We describe the datasets used, model implementation details, and baseline configurations. To systematically evaluate our method, we design experiments to answer the following three research questions:

\noindent \textbf{RQ1.} \textit{Does option weights improve LLMs' knowledge tracing performance in cold-start settings?}
We evaluate whether LOKT outperforms both traditional KT and LLM-based KT methods when cold-start settings. Additionally, we conduct experiments under warm-start settings to verify the robustness of LOKT.

\noindent \textbf{RQ2.} \textit{What are the scalability and efficiency benefits of using option-weight encapsulated prompts in LLM-based knowledge tracing?}
We conduct experiments comparing inference costs across various few-shot settings and evaluate performance under fixed token budgets to assess the impact of encapsulating option weights.

\noindent \textbf{RQ3.} \textit{To what extent does TCOW improve the knowledge tracing performance of LLMs?}
We assess whether TCOW improves LLM reasoning by comparing representation formats and textual semantic variants.


\subsection{Datasets and Baselines}

\noindent\textbf{\textit{Dataset.}} We used five public MCQ datasets: Eedi B, A \cite{Eedi}, EdNet \cite{ednet}, Assistment09, and DBE-KT22 \cite{dbe} (see Table~\ref{table1}). Eedi B, Eedi A, and EdNet have 4 options per question, while Assistment09 and DBE-KT22 have variable options (up to 7 and 5, respectively). We removed instances with missing or unselected answers during preprocessing.

\begin{table}[ht]

\centering
\caption{Datasets Description}
\label{table1}
  

\resizebox{\columnwidth}{!}{%
\begin{tabular}{l|ccccc}
\hline
 & \textbf{Eedi B} & \textbf{Eedi A} & \textbf{EdNet} & \textbf{Assistment09} & \textbf{DBE-KT22} \\ \hline
\textbf{\# Interactions} & 1,377,653 & 15,840,680 & 95,293,926 & 238,307 & 307,058 \\
\textbf{\# Learners}      & 4,918   & 118,971  & 784,309  & 7,353   & 1,264 \\
\textbf{\# Questions}     & 948     & 27,613   & 12,284   & 1,993   & 212 \\
\textbf{\# KCs}           & 86      & 388      & 188      & 315     & 93 \\
\textbf{Sparsity}         & 70.45\% & 99.52\%  & 99.01\%  & 98.37\% & 85.41\% \\
\textbf{\# Options}       & 4       & 4        & 4        & 7 (max) & 5 (max) \\
\hline
\end{tabular}%
}
\end{table}

\begin{table*}
\centering
\caption{Performance comparison of ACC and F1 scores (mean$\pm$std) at a training ratio of \( \rho_{\text{train}} = 0.001 \). \textbf{Bold} indicates the best performance, and \textit{underline} denotes the second-best.}
 
\label{tab:task1_performance}

\resizebox{\textwidth}{!}{
\begin{tabular}{l|cc|cc|cc|cc|cc}
\hline
\multicolumn{1}{c|}{\textbf{}} & \multicolumn{2}{c|}{\textbf{Eedi B}} & \multicolumn{2}{c|}{\textbf{Eedi A}} & \multicolumn{2}{c|}{\textbf{Ednet}} & \multicolumn{2}{c|}{\textbf{Assistment09}} & \multicolumn{2}{c}{\textbf{DBE-KT22}}  \\ \hline
\multicolumn{1}{c|}{\textbf{Model}} & \textbf{ACC} & \textbf{F1} & \textbf{ACC} & \textbf{F1} & \textbf{ACC} & \textbf{F1} & \textbf{ACC} & \textbf{F1} & \textbf{ACC} & \textbf{F1} \\ \hline

 \multicolumn{11}{l}{\textbf{Non-LLM Based}} \\
\hline
DKT    & 0.495{\scriptsize$\pm$0.044} & 0.496{\scriptsize$\pm$0.041} & 0.489{\scriptsize$\pm$0.061} & 0.497{\scriptsize$\pm$0.062} & 0.527{\scriptsize$\pm$0.050} & 0.513{\scriptsize$\pm$0.077} & 0.490{\scriptsize$\pm$0.067} & 0.495{\scriptsize$\pm$0.076} & 0.471{\scriptsize$\pm$0.048} & 0.481{\scriptsize$\pm$0.048} \\

DKVMN  & 0.515{\scriptsize$\pm$0.048} & 0.526{\scriptsize$\pm$0.034} & 0.503{\scriptsize$\pm$0.023} & 0.503{\scriptsize$\pm$0.034} & 0.539{\scriptsize$\pm$0.021} & 0.521{\scriptsize$\pm$0.060} & 0.459{\scriptsize$\pm$0.054} & 0.459{\scriptsize$\pm$0.071} & 0.490{\scriptsize$\pm$0.034} & 0.520{\scriptsize$\pm$0.035} \\
SAKT   & 0.493{\scriptsize$\pm$0.085} & 0.494{\scriptsize$\pm$0.092} & 0.513{\scriptsize$\pm$0.091} & 0.497{\scriptsize$\pm$0.093} & 0.483{\scriptsize$\pm$0.034} & 0.540{\scriptsize$\pm$0.092} & 0.483{\scriptsize$\pm$0.092} & 0.540{\scriptsize$\pm$0.099} & 0.548{\scriptsize$\pm$0.081} & 0.623{\scriptsize$\pm$0.083} \\
AKT    & 0.535{\scriptsize$\pm$0.065} & 0.476{\scriptsize$\pm$0.024} & 0.523{\scriptsize$\pm$0.045} & 0.538{\scriptsize$\pm$0.048} & 0.538{\scriptsize$\pm$0.023} & 0.645{\scriptsize$\pm$0.048} & 0.507{\scriptsize$\pm$0.084} & 0.574{\scriptsize$\pm$0.071} & 0.493{\scriptsize$\pm$0.028} & 0.659{\scriptsize$\pm$0.059} \\

ExtraKT& 0.573{\scriptsize$\pm$0.018} & 0.518{\scriptsize$\pm$0.022} & 0.561{\scriptsize$\pm$0.018} & 0.534{\scriptsize$\pm$0.020} & 0.552{\scriptsize$\pm$0.016} & 0.522{\scriptsize$\pm$0.021} & 0.568{\scriptsize$\pm$0.018} & 0.529{\scriptsize$\pm$0.023} & 0.571{\scriptsize$\pm$0.022} & 0.533{\scriptsize$\pm$0.023} \\ 
ReKT   & 0.577{\scriptsize$\pm$0.019} & 0.522{\scriptsize$\pm$0.023} & 0.563{\scriptsize$\pm$0.018} & 0.538{\scriptsize$\pm$0.021} & 0.554{\scriptsize$\pm$0.017} & 0.527{\scriptsize$\pm$0.019} & 0.571{\scriptsize$\pm$0.019} & 0.532{\scriptsize$\pm$0.024} & 0.573{\scriptsize$\pm$0.021} & 0.537{\scriptsize$\pm$0.024} \\
\hline

\multicolumn{11}{l}{\textbf{LLM-Based (GPT-4o)}} \\ \hline
\cite{neshaei2024towards} & 0.660{\scriptsize$\pm$0.015} & \underline{0.679{\scriptsize$\pm$0.014}} & 0.624{\scriptsize$\pm$0.019} & 0.615{\scriptsize$\pm$0.018} & \underline{0.581{\scriptsize$\pm$0.020}} & \textbf{0.656{\scriptsize$\pm$0.016}} & \underline{0.632{\scriptsize$\pm$0.015}} & \textbf{0.681{\scriptsize$\pm$0.009}} & 0.608{\scriptsize$\pm$0.017} & 0.688{\scriptsize$\pm$0.014} \\
CLST   & \underline{0.674{\scriptsize$\pm$0.007}} & 0.678{\scriptsize$\pm$0.011} & \underline{0.638{\scriptsize$\pm$0.015}} & \underline{0.660{\scriptsize$\pm$0.016}} & 0.577{\scriptsize$\pm$0.020} & 0.616{\scriptsize$\pm$0.018} & 0.620{\scriptsize$\pm$0.012} & 0.651{\scriptsize$\pm$0.017} & \underline{0.657{\scriptsize$\pm$0.019}} & \underline{0.693{\scriptsize$\pm$0.018}} \\ 
\rowcolor{gray!20}
LOKT  & \textbf{0.694{\scriptsize$\pm$0.017}} & \textbf{0.683{\scriptsize$\pm$0.019}} & \textbf{0.686{\scriptsize$\pm$0.015}} & \textbf{0.698{\scriptsize$\pm$0.017}} & \textbf{0.668{\scriptsize$\pm$0.012}} & \underline{0.646{\scriptsize$\pm$0.011}} & \textbf{0.640{\scriptsize$\pm$0.016}} & \underline{0.644{\scriptsize$\pm$0.016}} & \textbf{0.698{\scriptsize$\pm$0.011}} & \textbf{0.737{\scriptsize$\pm$0.019}} \\ \hline

\end{tabular}
}
\end{table*}

\begin{figure*}
    \centering
\includegraphics[width=1\linewidth]{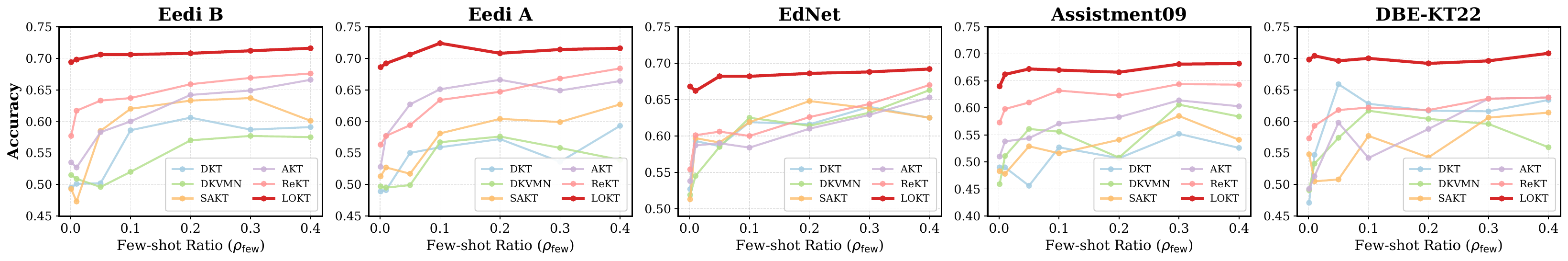}
\vspace{-8mm}
\caption{Performance comparison across different training ratios (\( \rho_{\text{few}} \) from 0.001 to 0.4). LLM-based baselines are excluded due to prompt length limitations in ICL. LOKT results are reported using GPT-4o.}
 
    \label{fewshot}
\end{figure*}

\noindent \textbf{\textit{Baseline Models.}}
We compare our model against state-of-the-art KT methods, including traditional approaches such as DKT \cite{Piech2015}, DKVMN \cite{zhang2017dynamic}, SAKT \cite{Xia2019}, AKT \cite{ghosh2020context}, ExtraKT \cite{li2024extending}, and ReKT \cite{shen2024revisiting}. In addition, we consider recent LLM-based methods like \cite{neshaei2024towards} and CLST \cite{jung2024}. Although these LLM-based models originally involve fine-tuning, their primary contribution lies in novel prompt design. Therefore, we adopt their prompt formats and evaluate all models in an inference-only setting to ensure a fair comparison with our approach. Detailed descriptions of the baseline models are provided in the Appendix \ref{baseline}.

\subsection{Model Training Setup}

All experiments use five different random seeds. Each dataset are split into 70\% training, 15\% validation, and 15\% test sets. Non-LLM models use 100-dimensional embeddings and are trained with the Adam optimizer, with both the learning rate and weight decay set to 0.001. A batch size of 256 is used, and early stopping is applied with a patience of 5 epochs. The training data is sampled per label according to the given few-shot ratio (\(\rho_{\text{few}}\)), and each learner's sequence length is limited to 50. All training runs on a system equipped with an NVIDIA RTX 3090 Ti GPU.
 
\subsection{Inference and Evaluation Protocol}
 
Both LLM-based and non-LLM models are evaluated using a balanced test set consisting of 50 correct and 50 wrong responses. This setting controls API usage and ensures fair comparisons across models. LLM baselines operate in an inference-only mode via the OpenAI API, primarily using few-shot ICL prompts, with ID-based prompts applied. Our proposed method, LOKT, is evaluated in a without in-context examples input directly rather than use uncapsulted . All reported results are averaged over multiple random seeds.

\section{Results}
  
\subsection{Performance in Cold-Start (RQ1)}
    
\noindent\textbf{\textit{Extreme Cold-Start Results.}} Table~\ref{tab:task1_performance} presents accuracy and F1 scores under an extreme cold-start setting (\(\rho_{\text{few}} = 0.001\)) across five public datasets, where LLM-based models consistently outperform traditional KT approaches, demonstrating superior generalization with minimal data. Notably, \textbf{LOKT} achieves the best overall performance by leveraging structured, context-aware prompts and incorporating TCOW, enabling more accurate predictions of learner responses. Compared to prior LLM-based baselines such as CLST and~\cite{neshaei2024towards}, LOKT provides richer semantic representations, maintaining robust performance even under severe data scarcity.

\noindent\textbf{\textit{Performance across Few-shot Ratios.}} Figure~\ref{fewshot} illustrates the performance of various KT models, under different Few-shot ratios. Overall, LOKT consistently achieves the highest accuracy across all datasets. This strong performance is attributed to LOKT’s use of TCOW, which quantify not only correct answers but also the degree of understanding and misconceptions reflected in wrong choices. LLM-based ICL models such as CLST and \cite{neshaei2024towards} are excluded due to prompt length limitations. This indicates that while ICL methods can partially address the cold-start problem, they face challenges in scalability and efficiency. In this regard, LOKT presents a more practical and robust solution for cold-start KT.

\noindent\textbf{\textit{Evaluating Generalization Capacity.}}

\begin{figure}[ht]
    \centering
    \includegraphics[width=1\linewidth]{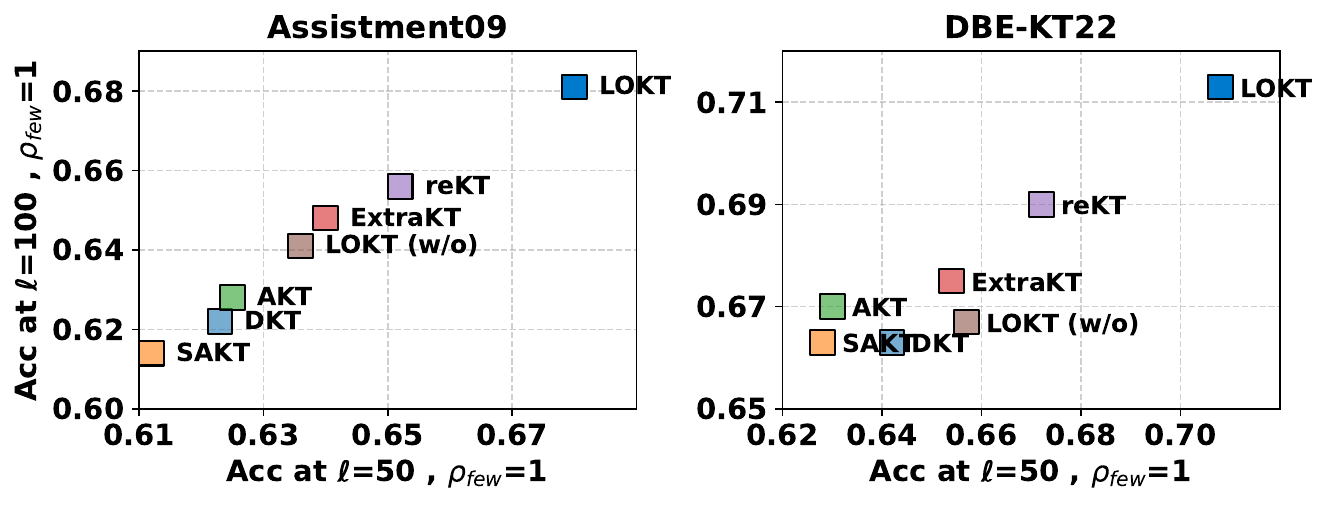}
    \vspace{-8mm}
    \caption{Accuracy at interaction sequence lengths \(\ell=50\) and \(\ell=100\) with full training data (\(\rho_{\text{few}}=1\)) on Assistment09 and DBE-KT22 datasets. LOKT performs best, particularly with longer sequences.}
    \label{warmup}
    
\end{figure}

 We evaluate the generalization performance of incorporating option weights under warm-start settings (\(\rho_{\text{few}} = 1\)) to complement our main focus on cold-start scenarios. Experiments were conducted using two interaction lengths, 50 and 100. As shown in Figure~\ref{warmup}, LOKT consistently outperformed baseline models, with a significant performance gain observed when using option weights compared to without option weights (LOKT w/o wgt). These results demonstrate that option-weighted representations enable LOKT to effectively capture nuanced learner behaviors across various data availability conditions, highlighting its robustness and practical applicability beyond cold-start settings.

\subsection{Scalability and Efficiency (RQ2)}

To evaluate the scalability and efficiency of the proposed encapsulated format using option weights, we compare LOKT with an ICL-based baseline. The baseline suffers from increased prompt lengths as the number of examples grows, leading to higher inference costs and potential context limit overflows. This study shows whether LOKT can overcome these challenges and maintain effective performance under resource-constrained conditions. To this end, we design two independent experiments: one focusing on prompt scalability and the other on performance under a fixed token budget.

\begin{figure}[ht]
\centering
\includegraphics[width=1\linewidth]{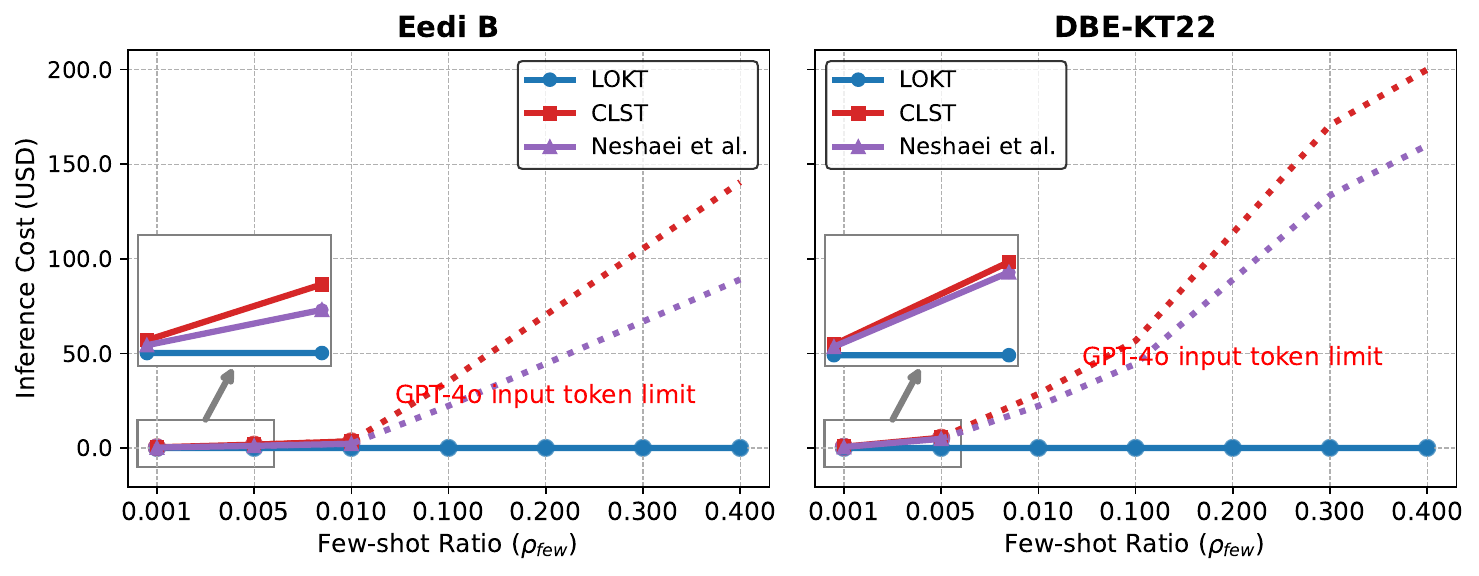}

\caption{Comparison of inference cost (USD) across training ratios. LOKT keeps token usage low and stable, while ICL-based methods incur higher costs and may exceed GPT-4o’s context limit.}
\label{fig:enter-Efficiency}

\end{figure}

\noindent\textbf{\textit{Prompt Length and Token Cost.}} As shown in Figure~\ref{fig:enter-Efficiency}, token usage in ICL increases linearly with the number of few-shot examples, resulting in higher inference costs and potential prompt overflow. In contrast, LOKT compresses information through TCOW, maintaining a nearly constant prompt length. These results highlight LOKT’s superior scalability and token-level efficiency. The inference cost was computed based on GPT-4o pricing at \$3.75\footnote{This price is based on the OpenAI API pricing as of April 28, 2025.} per million input tokens, measured per single test input. Total cost includes input and output tokens per query, emphasizing the economic advantage of LOKT’s compact prompts.

\begin{figure}[ht]
\centering
\includegraphics[width=0.7\linewidth]{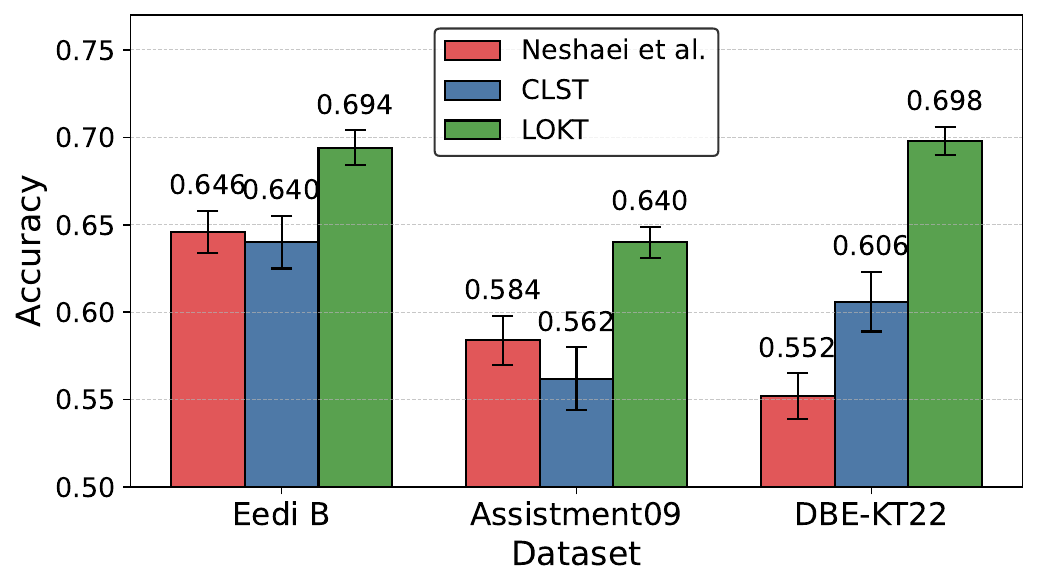}
   \vspace{-4mm}
\caption{Accuracy under a fixed token budget (2000 tokens) on Eedi B, Assistment09, and DBE-KT22.}
\label{fig:enter-Resource}

\end{figure}

\noindent\textbf{\textit{Performance under Resource Constraints.}} We fix the token budget to a maximum of 2000 tokens, reflecting LOKT’s typical prompt length, and compare model performance under identical conditions. As shown in Figure\ref{fig:enter-Resource}, LOKT consistently achieves the highest accuracy across the three datasets. This demonstrates that CTOW-based prompts effectively capture learners' partial understanding and misconceptions. In contrast, CLST and Neshaei et al.~\cite{neshaei2024towards} exhibit reduced or unstable performance with limited resources in cold-start scenarios. These findings underscore LOKT’s strength in balancing accuracy and resource efficiency.

\subsection{Effect of TCOW on KT Performance of LLMs (RQ3)}
 
To verify whether our proposed option weight method, TCOW, effectively compresses learners’ problem-solving histories while preserving rich information, we compare it against various alternative option weighting methods. The first experiment aims to demonstrate the advantages of semantically rich textual representations over continuous and ordinal formats. The second experiment examines the impact of preserving meaningful semantic structure on model performance by comparing TCOW with coarse-grained and randomly shuffled variants.




\noindent\textbf{\textit{Effect of Option Weight Representation.}}
Table~\ref{ft_fewshot} presents a comparison of three option weight representations: continuous, ordinal, and TCOW. This experiment shows the effectiveness of semantically strutured prompts in LLMs' knowledge tracing. The results demonstrate that TCOW consistently outperforms both continuous and ordinal across all datasets. TCOW integrates pedagogically aligned semantic categories that support an LLM reason more effectively with regard to learners' understanding, leading to clear performance gains. While continuous provides a highly compact encoding and achieves comparable performance to standard ICL-based models, it lacks sufficient structure for LLMs to interpret nuanced learner behavior. Ordinal improves slightly by incorporating ranking information, but the absence of semantic content limits its effectiveness. This confirms that semantic structure, not just compression, is key to enabling LLMs in knowledge tracing.
\begin{table}[!htt]
\centering
\setlength{\tabcolsep}{3pt}
\caption{Performance with different option weight formats under GPT-4o at \(\rho_{\text{few}} = 0.001\).}
\label{ft_fewshot}
 
\resizebox{\columnwidth}{!}{
\begin{tabular}{l|cc|cc|cc|cc|cc}
\hline
\textbf{Method} & \multicolumn{2}{c|}{\textbf{Eedi B}} & \multicolumn{2}{c|}{\textbf{Eedi A}} & \multicolumn{2}{c|}{\textbf{EdNet}} & \multicolumn{2}{c|}{\textbf{Assistment09}} & \multicolumn{2}{c}{\textbf{DBE-KT22}} \\
\cline{2-11}
& \textbf{ACC} & \textbf{F1} & \textbf{ACC} & \textbf{F1} & \textbf{ACC} & \textbf{F1} & \textbf{ACC} & \textbf{F1} & \textbf{ACC} & \textbf{F1} \\
\hline
Continuous & 0.656 & \underline{0.625} & 0.672 & 0.694 & 0.622 & 0.505 & 0.620 & 0.614 & 0.658 & 0.707 \\
Ordinal & \underline{0.668} & 0.617 & \underline{0.676} & \underline{0.696} & \underline{0.630} & \underline{0.525} & \underline{0.630} & \underline{0.627} & \underline{0.670} & \underline{0.714} \\
\rowcolor{gray!20}
TCOW & \textbf{0.694} & \textbf{0.683} & \textbf{0.686} & \textbf{0.698} & \textbf{0.668} & \textbf{0.640} & \textbf{0.640} & \textbf{0.644} & \textbf{0.698} & \textbf{0.737} \\
\hline
\end{tabular}
}
\end{table}

\noindent\textbf{\textit{Effect of Semantic Textual Alignment.}  }
To evaluate the impact of semantic alignment in textual categorical option weights, we compare three design strategies. The first variant, \textit{Random}, retains the categorical labels (e.g., \textit{Proficient}, \textit{Partial}, \textit{Limited}, \textit{Inadequate}) but randomly shuffles their assignment to options, disregarding correctness and semantic structure. The second, \textit{Binary}, simplifies the labels into two coarse-grained categories: \textit{Proficient} and \textit{Limited}, eliminating intermediate distinctions. The final variant, \textbf{TCOW} (our proposed method), preserves the hierarchy and ensures that label assignment reflects the semantic order implied by the option rankings. As shown in Table~\ref{fig:combined-label}, TCOW significantly outperforms the other variants, demonstrating that maintaining semantic alignment in textual prompts leads to more accurate modeling of learner understanding. For another few-shot settings, please refer to the Appendix~F.3.

\begin{table}[ht]
\centering
\caption{Performance at \(\rho_{\text{few}} = 0.001\) using TCOW with varying semantic textual alignment under GPT-4o.}
\label{fig:combined-label}
\setlength{\tabcolsep}{2pt}
 
\resizebox{\columnwidth}{!}{
\begin{tabular}{l|cc|cc|cc|cc|cc}
\hline
\textbf{}  & \multicolumn{2}{c|}{\textbf{Eedi B}} & \multicolumn{2}{c|}{\textbf{Eedi A}} & \multicolumn{2}{c|}{\textbf{Ednet}} & \multicolumn{2}{c|}{\textbf{Assistment09}} & \multicolumn{2}{c}{\textbf{DBE-KT22}} \\
\hline
\textbf{Model} & \textbf{ACC} & \textbf{F1} & \textbf{ACC} & \textbf{F1} & \textbf{ACC} & \textbf{F1} & \textbf{ACC} & \textbf{F1} & \textbf{ACC} & \textbf{F1} \\
\hline
Random    & 0.662 & {0.612} & 0.656 & 0.648 & 0.593 & {0.560} & \underline{0.630} & 0.580 & 0.654 & 0.658 \\
Binary    & \underline{0.666} &   \underline{0.659} & \underline{0.656} & \underline{0.651} & \underline{0.632} &  \underline{0.648} & 0.620 & \underline{0.619} & \underline{0.659} & \underline{0.684} \\
\rowcolor{gray!20}
TCOW      & \textbf{0.694} & \textbf{0.683} & \textbf{0.686} & \textbf{0.698} & \textbf{0.668} & \textbf{0.640} & \textbf{0.640} & \textbf{0.644} & \textbf{0.698} & \textbf{0.737} \\
\hline
\end{tabular}
}
\end{table}

\subsection{Further Analysis}
To further examine the impact of incorporating option weights into a prompt-template for LLM-based KT, we conducted additional experiments, including an ablation study, model size analysis.

\noindent\textbf{\textit{Ablation of Option Weight.}} In Table~\ref{prompt_mode}, LOKT (w/o opt, wgt) removes both option and option weight, LOKT (w/o wgt) omits only option weight. Including only the option sequence yields some improvement, but the best results are achieved when both the option sequence and weights are incorporated. In particular, using TCOW allows the LLM to more accurately interpret the diagnostic meaning of each choice and the learner’s understanding. This enables LOKT to capture partial knowledge and misconceptions more effectively than previous methods, highlighting the importance of semantically enriching the prompt for superior performance.

\begin{table}[!htt]
\centering
\caption{Performance for \(\rho_{\text{few}} = 0.001\) using GPT-4o.  
O = Option Prompt, OW = Option Weight}
\label{prompt_mode}
  
\setlength{\tabcolsep}{2pt}
\resizebox{\columnwidth}{!}{
\begin{tabular}{cc|cc|cc|cc|cc|cc}
\hline
\textbf{O} & \textbf{OW} & \multicolumn{2}{c|}{\textbf{Eedi B}} & \multicolumn{2}{c|}{\textbf{Eedi A}} & \multicolumn{2}{c|}{\textbf{EdNet}} & \multicolumn{2}{c|}{\textbf{Assistment09}} & \multicolumn{2}{c}{\textbf{DBE-KT22}} \\
\cline{3-12}
& & \textbf{ACC} & \textbf{F1} & \textbf{ACC} & \textbf{F1} & \textbf{ACC} & \textbf{F1} & \textbf{ACC} & \textbf{F1} & \textbf{ACC} & \textbf{F1} \\
\hline
\multirow{1}{*}{\centering\xmark} & \multirow{1}{*}{\centering\xmark} 
& 0.660 & 0.628 & 0.660 & 0.691 & 0.580 & 0.609 & \underline{0.636} & \textbf{0.658} & 0.666 & \underline{0.706} \\
\multirow{1}{*}{\centering\cmark} & \multirow{1}{*}{\centering\xmark} 
& \underline{0.667} & \underline{0.641} & \underline{0.664} & \underline{0.694} & \underline{0.609} & \underline{0.630} & 0.634 & 0.632 & \underline{0.668} & 0.705 \\
\rowcolor{gray!20}
\multirow{1}{*}{\centering\cmark} & \multirow{1}{*}{\centering\cmark} 
& \textbf{0.694} & \textbf{0.683} & \textbf{0.686} & \textbf{0.698} & \textbf{0.668} & \textbf{0.640} & \textbf{0.640} & \underline{0.644} & \textbf{0.698} & \textbf{0.737} \\
\hline
\end{tabular}
}  
\end{table}

\begin{figure}[ht]
\centering

    \includegraphics[width=1\linewidth]{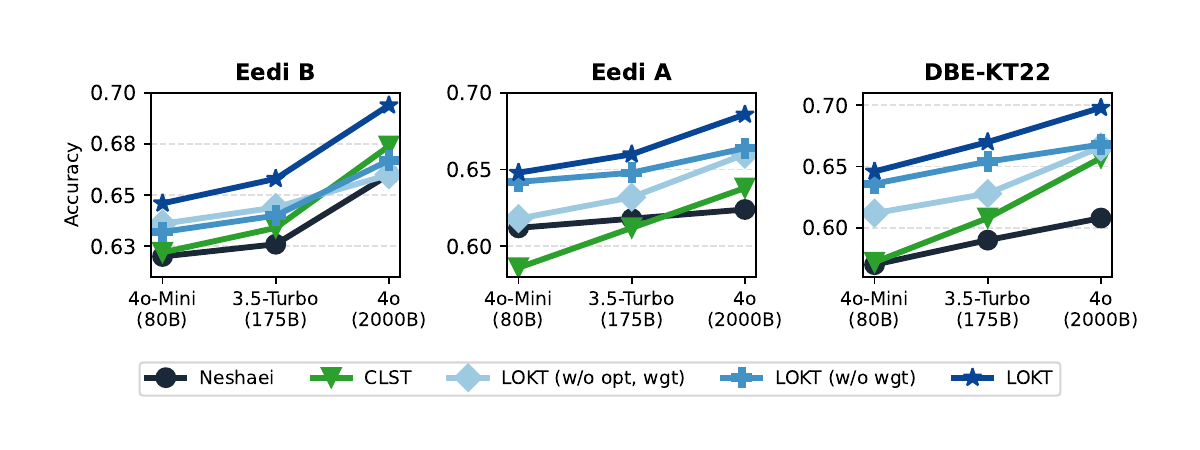}
    \vspace{-8mm}
    \caption{Accuracy of LOKT and baselines across GPT sizes (4o-Mini: 80B, 3.5-Turbo: 175B, 4o: 2000B).}
    \label{fig:model_size}
\end{figure}

\noindent\textbf{\textit{Performance by LLM Parameter Size.}} LOKT consistently outperforms baseline methods across all model sizes, as shown in Figure~\ref{fig:model_size}. As the LLM size increases, the performance gap in favor of LOKT becomes more pronounced, indicating that larger models benefit even more from the option weight-based prompting scheme. These results highlight the robustness and scalability of LOKT across different LLM environments.
 \vspace{-1mm}

 \vspace{-3mm}
\section{Conclusion}
 \vspace{-1mm}
In this paper, we analyze the fundamental limitations of existing methods in LLM-based knowledge tracing models developed to tackle the cold-start problem, focusing on inefficiency and scalability issues caused by long interaction histories. To address these challenges, we propose LOKT, a simple yet effective framework that compresses learner interactions into textual categorical option weights within prompts, enabling large language models to efficiently understand learners’ problem-solving histories. Extensive experiments on multiple public datasets demonstrate that LOKT surpasses conventional KT models and LLM-based in-context learning methods in accuracy, scalability, and cost efficiency. Our study highlights that leveraging effective compression techniques in in-context learning is key to scalable and practical knowledge tracing with LLMs.

\section*{Limitations}
 \vspace{-1mm}
This paper has two main limitations. First, it lacks human evaluation of LLM predictions, limiting qualitative validation of interpretability and real-world applicability. Second, the option weight calculation depends on statistical assumptions that may not fully match the data, potentially affecting performance. Future work will include expert assessments and broader experiments to address these issues.

\bibliography{custom}

\clearpage

\appendix

\section{Data Source}
 \vspace{-4mm}
\begin{table}[ht]
\centering
\caption{Dataset sources used in this study}
\small
\begin{tabularx}{\linewidth}{l|X}
\toprule
\textbf{Dataset} & \textbf{Source URL} \\
\midrule
Eedi B & \url{https://Eedi.com/research} \\
Eedi A & \url{https://Eedi.com/research} \\
EdNet & \url{https://github.com/riiid/ednet} \\
Assistment09 & \url{https://sites.google.com/site/assistmentsdata} \\
DBE-KT22 & \url{https://doi.org/10.26193/6DZWOH} \\
\bottomrule
\end{tabularx}

\label{tab:dataset_sources}
\end{table}

 \vspace{-4mm}
\section{Baselines}
\label{baseline}
\begin{itemize}
  \item \textbf{DKT}: A recurrent neural network model that captures learners' knowledge states over time based on their interaction sequences.
  \item \textbf{DKVMN}: A dynamic key-value memory network that models individual knowledge components with an external memory structure.
  \item \textbf{SAKT}: A self-attention based model that focuses on important past interactions to predict learner performance.
  \item \textbf{AKT}: An attentive knowledge tracing model incorporating question difficulty and knowledge component relations for improved predictions.
\item \textbf{ExtraKT}: A model applying linearly decayed attention bias to capture short-term forgetting, enabling robust performance across varying context lengths.
  \item \textbf{ReKT}: ReKT models student knowledge using questions, concepts, and domains with a lightweight cognitive-inspired FRU mechanism, achieving accurate KT with low computational cost.
  \item \textbf{(Neshaei et al., 2024)}:This study explores using pre-trained large language models for knowledge tracing, finding that while zero-shot approaches underperform, fine-tuned LLMs outperform naive baselines and match or exceed traditional Bayesian methods in predicting student performance.
  \item \textbf{CLST}: This study proposes CLST, a generative LLM-based framework for mitigating cold-start issues in knowledge tracing by representing problem-solving data in natural language and fine-tuning the model for effective cross-domain student performance prediction.
\end{itemize}

\section{Option Weight Assignment Method}\label{option_weight_assignment}

\subsection{Option Weight Representation Transformations}\label{option_weight_textual}

\begin{table}[ht]
    \centering
    \caption{Comparison of Transformation Methods: Continuous, Ordinal, and TCOW Assignments}
    \label{tab:answer_types}
    \resizebox{\columnwidth}{!}{%
        \begin{tabular}{l|p{0.2\columnwidth}|p{0.2\columnwidth}|p{0.2\columnwidth}|p{0.2\columnwidth}}
            \hline
            \textbf{Method} & \textbf{Correct Option} & \textbf{Wrong Option 1} & \textbf{Wrong Option 2} & \textbf{Wrong Option 3} \\
            \hline
            Continuous    & 0.3       & 0.1       & -0.2      & -0.5       \\
            \hline
            Ordinal       & 1         & 2         & 3         & 4          \\
            \hline
            \rowcolor[gray]{0.9}
            TCOW   & Proficient & Partial   & Limited   & Inadequate \\
            \hline
        \end{tabular}
    }
\end{table}

\noindent This section explains how continuous option weights, such as those shown in Table~\ref{tab:answer_types} (e.g., 0.3, 0.1, -0.2, -0.5), which can be challenging for large language models (LLMs) to interpret directly, are transformed into more interpretable formats. The ordinal transformation assigns a rank to each option based on its weight (e.g., 1, 2, 3, 4), while the categorical transformation maps these weights into meaningful textual labels such as \textit{Proficient}, \textit{Partial}, \textit{Limited}, and \textit{Inadequate}. By converting numerical weights into semantically rich categories, as shown in the categorical row of Table~\ref{tab:answer_types}, this approach better leverages the strengths of LLMs and enhances the accuracy of knowledge tracing.

\subsection{Categorical Option Weight Semantic Transformations}\label{option_weight_finegrained}
\begin{table}[httbp]
    \centering
    \caption{Comparison of Transformation Methods: Random, Binary, and TCOW Assignments}

    \resizebox{\columnwidth}{!}{%
        \begin{tabular}{l|p{0.2\columnwidth}|p{0.2\columnwidth}|p{0.2\columnwidth}|p{0.2\columnwidth}}
            \hline
            \textbf{Method} & \textbf{Correct Option} & \textbf{Wrong Option 1} & \textbf{Wrong Option 2} & \textbf{Wrong Option 3} \\
            \hline
            Random       & Inadequate & Partial  & Limited  & Proficient \\
            \hline
            Binary       & Proficient & Inadequate & Inadequate & Inadequate \\
            \hline
            \rowcolor[gray]{0.9}
            TCOW & Proficient & Partial  & Limited  & Inadequate \\
            \hline
        \end{tabular}
    }
    \label{tab:answer_types2}
\end{table}

\noindent This section compares three methods for transforming option weights into categorical representations: \textit{Random}, \textit{Binary}, and \textit{Textual Categorical Option Weight (TCOW)}. As shown in Table~\ref{tab:answer_types2}, random transformations assign labels arbitrarily, which leads to inconsistent performance. Binary transformations classify the correct answer as \textit{Proficient} and all wrong answers uniformly as \textit{Inadequate}, but this fails to distinguish finer differences among wrong options. In contrast, the fine-grained TCOW method assigns more detailed categories such as \textit{Partial} and \textit{Limited}, effectively capturing varying levels of learner understanding and misconceptions. This approach provides the best alignment for knowledge tracing tasks.



\begin{table*}[ht]
\centering
\small
\caption{Categorical Assignments Based on Number of Options}
\label{tab:option_weights}
\begin{tabularx}{\textwidth}{c >{\centering\arraybackslash}X *{6}{>{\centering\arraybackslash}X}}
\toprule
\textbf{\# Options} & \textbf{Correct} & \multicolumn{6}{c}{\textbf{Wrong}} \\
\cmidrule(lr){2-2} \cmidrule(lr){3-8}
 & 1st & 2nd & 3rd & 4th & 5th & 6th & 7th \\
\midrule
2 & \cellcolor{green!20}Proficient & \cellcolor{red!15}Inadequate & & & & & \\
3 & \cellcolor{green!20}Proficient & \cellcolor{red!15}Limited & \cellcolor{red!15}Inadequate & & & & \\
4 & \cellcolor{green!20}Proficient & \cellcolor{red!15}Partial & \cellcolor{red!15}Limited & \cellcolor{red!15}Inadequate & & & \\
5 & \cellcolor{green!20}Proficient & \cellcolor{red!15}Partial & \cellcolor{red!15}Limited & \cellcolor{red!15}Inadequate & \cellcolor{red!15}Inadequate & & \\
6 & \cellcolor{green!20}Proficient & \cellcolor{red!15}Partial & \cellcolor{red!15}Limited & \cellcolor{red!15}Limited & \cellcolor{red!15}Inadequate & \cellcolor{red!15}Inadequate & \\
7 & \cellcolor{green!20}Proficient & \cellcolor{red!15}Partial & \cellcolor{red!15}Partial & \cellcolor{red!15}Limited & \cellcolor{red!15}Limited & \cellcolor{red!15}Inadequate & \cellcolor{red!15}Inadequate \\
\bottomrule
\end{tabularx}
\end{table*}

\begin{figure*}[ht]
    \centering
    \begin{subfigure}{\linewidth}
        \centering
                \includegraphics[width=\linewidth]{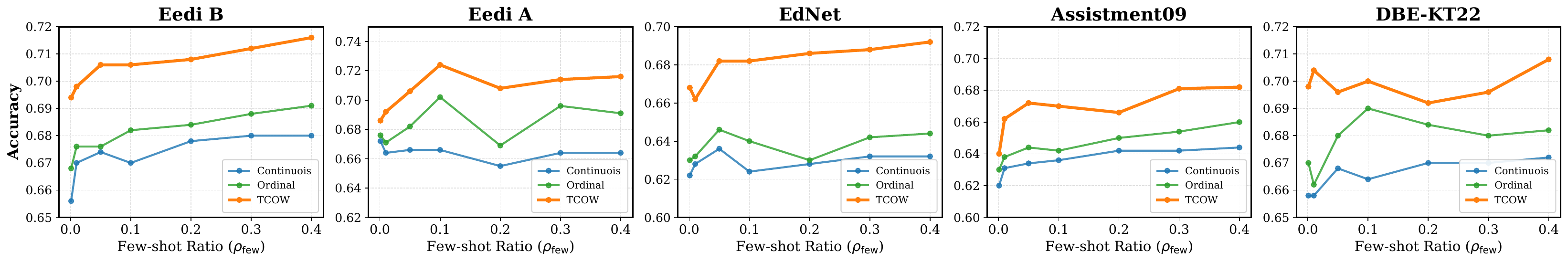}
        \caption{Performance metrics for different representation configurations, comparing Continuous, Ordinal, and TCOW (our proposed).}
        \label{fig:option_weight_representation}

    \end{subfigure}
    \vspace{1mm}
    \begin{subfigure}{\linewidth}
        \centering
        \includegraphics[width=\linewidth]{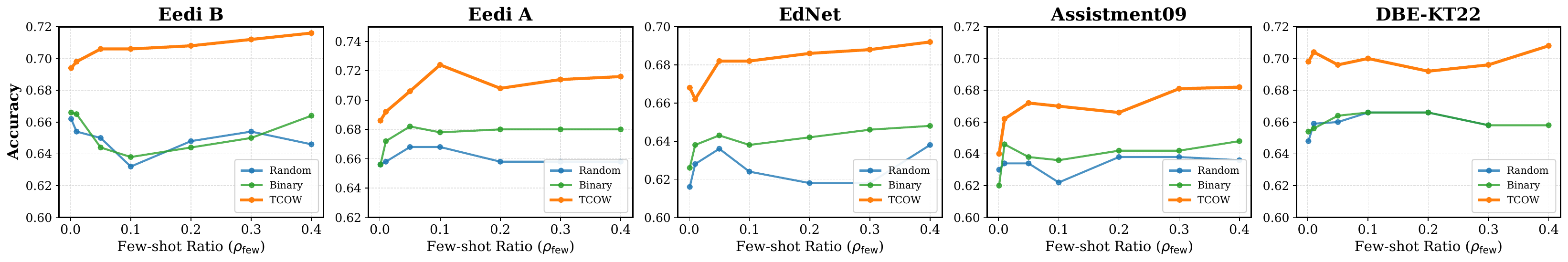}
        \caption{Performance metrics for different semantic configurations, comparing Random, Binary, and TCOW (our proposed).}
        \label{fig:option_weight_semantic}
    \end{subfigure}
     \vspace{-5mm}
    \caption{
        Ablation analysis for option weight strategies.
        (a) Semantic configuration variants.
        (b) Representation format variants.
    }
    \vspace{-5mm}
    \label{fig:option_weight_ablation}
\end{figure*}

\subsection{Option Weight Expansion for MCQ}
\label{multiple_choice}
This section presents the categorizing scheme for option weights in multiple-choice questions (MCQs) based on the number of options. For MCQs with four options, TCOW, namely \textit{Proficient}, \textit{Partial}, \textit{Limited}, and \textit{Inadequate}, are assigned for each option in descending order. For MCQs with more than four options, additional TCOWs are allocated to the extra options in ascending order of weight values, with a maximum of two. This allocation aims to balance the number of TCOWs while placing greater emphasis on options with higher weights, based on the assumption that these weights carry richer information.

\section{Performance in Diverse Few-shot}

\subsection{Effect of Option Weight Representation}
\label{representation}
As shown in Figure \ref{fig:option_weight_representation}, text-based categorical weights consistently exhibit superior performance across various datasets and scenarios. They perform particularly well in shorter interaction sequences and maintain robust accuracy as sequence lengths increase. The figure highlights that text-based categorical weights deliver consistent results even in data-limited environments, aligning effectively with LLM input structures to enhance both predictive accuracy and interpretability. In contrast, continuous weights tend to struggle in sparse data conditions, failing to capture consistent patterns, while ordinal weights show some improvement but lack the contextual clarity of text-based weights.

\subsection{Effect of Textual Option Weight Semantic}
\label{semantic}

The figure \ref{fig:option_weight_semantic} shows that LOKT with fine-grained categorical weights consistently outperforms other weight methods (Random, Binary) across diverse few-shot settings. Even at low data ratios and short sequences, LOKT maintains high accuracy, with performance improving further as data ratios and sequence lengths increase. Unlike Binary weights, which fail to capture subtle differences in knowledge states, fine-grained categorical weights effectively leverage the hierarchical structure of options, aligning with LLMs' strength in processing semantically rich inputs. This demonstrates that the performance gains stem from meaningful semantic representations rather than merely converting weights into text.

\onecolumn

\section{Algorithm of option weight calculation}
\label{optionweight}
\begin{algorithm*}
\caption{Option Weights Calculation}

\begin{algorithmic}[1]
\REQUIRE{All learner records $X = \{ X_1, X_2, \dots, X_{|I|} \}$ with $|I|$ learners, $|Q|$ questions, and each question has $|O|$ options;  
A set of learners' proficiency  $\bar{x} = \{ \bar{x}_1, \bar{x}_2, \ldots, \bar{x}_{|I|} \}$;  
Standard deviation of average proficiency of all learners $S_x$;}
\ENSURE{The weight matrix $W = [w^o_q]_{|Q| \times |O|}$}

\STATE // Calculating the standard deviation of learner score
\STATE $\bar{x} \gets \frac{i}{|I|} \sum_{i \in I} \bar{x}_l$ \hfill // Average score calculating
\STATE $S_x \gets \sqrt{ \frac{1}{|I|} \sum_{l \in I} (\bar{x}_i - \bar{x})^2 }$ \hfill // Standard deviation calculating

\STATE // Calculating the proportion of selected and unselected for each option
\FOR{$q = 1, 2, \dots, |Q|$}
    \FOR{$o = 1, 2, \dots, |O|$}
        \STATE $c^o_q \gets$ number of learners who selected option $o$ on question $q$;
        \STATE $nc^o_q \gets$ number of learners who did \emph{not} select option $o$ on question $q$;
        \STATE $C^o_q \gets \frac{c^o_q}{c^o_q + nc^o_q}$;
        \STATE $NC^o_q \gets \frac{nc^o_q}{c^o_q + nc^o_q}$;
    \ENDFOR
\ENDFOR

\STATE // Calculating the average score of learners who selected or unselected each option
\FOR{$q = 1, 2, \dots, |Q|$}
    \FOR{$o = 1, 2, \dots, |O|$}
        \STATE $I^o_q \gets$ Set of learners who selected option $o$ on question $q$;
        \STATE $NI^o_q \gets$ Set of learners who did not select option $o$ on question $q$;
        \STATE $\bar{x}^o_{c_q} \gets \frac{1}{c^o_q} \sum_{i \in I^o_q} \bar{x}_i$;
        \STATE $\bar{x}^o_{nc_q} \gets \frac{1}{nc^o_q} \sum_{i \in NI^o_q} \bar{x}_i$;
    \ENDFOR
\ENDFOR

\STATE // Calculating the weight of each option for each question
\FOR{$q = 1, 2, \dots, |Q|$}
    \FOR{$o = 1, 2, \dots, |O|$}
        \STATE $w^o_q \gets \frac{ \bar{x}^o_{c_q} - \bar{x}^o_{nc_q} }{ S_x } \cdot \sqrt{ C^o_q \cdot NC^o_q }$;
    \ENDFOR
\ENDFOR

\RETURN $W$
\end{algorithmic}
\end{algorithm*}

\newpage

\lstset{
  basicstyle=\ttfamily\footnotesize,
  breaklines=true,
  frame=single,
  columns=fullflexible,
  keepspaces=true,
  showstringspaces=false,
  backgroundcolor=\color{gray!10},
  rulecolor=\color{black!50},
  xleftmargin=0pt,
  xrightmargin=0pt
}

\section{Example of System Messages and Predefined Template for LOKT }\label{prompt_1}
\subsection{LOKT}

\subsubsection{System Message}
\begin{lstlisting}
You are an advanced knowledge tracing expert. You track and predict the learner's knowledge states by considering their problem-solving history, associated Knowledge Components (KCs), selected options, and the weights of those selected options.
\end{lstlisting}

\subsubsection{Predefined Template}
\begin{lstlisting}
Analyze the learner's problem-solving history and predict their performance on the next question.
learner's problem-solving history:
- Question ID sequence: {question_ids}
- KC ID sequence: {kc_ids} (Note: This is a list of KC IDs associated with the next question. For example, if next_kc_id is [3, 72], it means the next question involves KC IDs 3 and 72.)
- Selected option sequence: {option_sequence}
- Selected option weights: {option_weights}
- Correctness sequence: {answer_sequence}

Next question details:
- Next question ID: {next_question_id}
- Next question's KC ID: {next_kc_id}

Based on the above information, predict whether the learner will answer the next question (ID: {next_question_id}, KC ID: {next_kc_id}) correctly ('correct') or wrongly ('wrong').

Consider the following when making your prediction:
- The learner's overall correctness pattern.
- The complexity and difficulty levels of the questions and KC IDs.
- How the selected options reflect their weight on the learner's understanding and confidence.
- Recent trends in the learner's performance.
- The learner's progression and knowledge improvement over time.
- The learner's current knowledge state for each KC and how it matches the KC IDs in the next question.
- Ignore any NaN values in the option weights.

Output only the single word ['correct' or 'wrong']. No other words or punctuation should be included.
\end{lstlisting}

\subsection{LOKT without option and option weight}
\subsubsection{System Message}
\begin{lstlisting}
You are an advanced knowledge tracing expert. You track and predict knowledge states by considering the learner's problem-solving history, KCs.
\end{lstlisting}

\subsubsection{Predefined Template}
\begin{lstlisting}
Analyze the learner's problem-solving history and predict their performance on the next question.
learner's problem-solving history:
- Question ID sequence: {question_ids}
- KC ID sequence: {kc_ids} (Note: This is a list of KC IDs associated with the next question. For example, if next_kc_id is [3, 72], it means the next question involves KC IDs 3 and 72.)
- Correctness sequence: {answer_sequence}

Next question details:
- Next question ID: {next_question_id}
- Next question's KC ID: {next_kc_id}

Based on the above information, predict whether the learner will answer the next question (ID: {next_question_id}, KC ID: {next_kc_id}) correctly ('correct') or wrongly ('wrong').

Consider the following when making your prediction:
- The learner's overall correctness pattern.
- The complexity and difficulty levels of the questions and KC IDs.
- Recent trends in the learner's performance.
- The learner's progression and knowledge improvement over time.
- The learner's current knowledge state for each KC and how it matches the KC IDs in the next question.

Provide a detailed assessment of how these factors influence the learner's knowledge state and the likelihood of correctly answering the next question.

Output only the single word ['correct' or 'wrong']. No other words or punctuation should be included.
\end{lstlisting}

\subsection{LOKT without option weight}

\subsubsection{System Message}
\begin{lstlisting}
You are an advanced knowledge tracing expert. You track and predict knowledge states by considering the learner's problem-solving history, selected options, KCs.
\end{lstlisting}

\subsubsection{Predefined Template}
\begin{lstlisting}
Analyze the learner's problem-solving history and predict their performance on the next question.
learner's problem-solving history:
- Question ID sequence: {question_ids}
- KC ID sequence: {kc_ids} (Note: This is a list of KC IDs associated with the next question. For example, if next_kc_id is [3, 72], it means the next question involves KC IDs 3 and 72.)
- Selected option sequence: {option_sequence}
- Correctness sequence: {answer_sequence}

Next question details:
- Next question ID: {next_question_id}
- Next question's KC ID: {next_kc_id}

Based on the above information, predict whether the learner will answer the next question (ID: {next_question_id}, KC ID: {next_kc_id}) correctly ('correct') or wrongly ('wrong').

Consider the following when making your prediction:
- The learner's overall correctness pattern.
- The complexity and difficulty levels of the questions and KC IDs.
- How the selected options reflect their weight on the learner's understanding and confidence.
- Recent trends in the learner's performance.
- The learner's progression and knowledge improvement over time.
- The learner's current knowledge state for each KC and how it matches the KC IDs in the next question.

Output only the single word ['correct' or 'wrong']. No other words or punctuation should be included.
\end{lstlisting}

\end{document}